\def\year{2021}\relax
\documentclass[letterpaper]{article} 
\usepackage{aaai21}  
\usepackage{times}  
\usepackage{helvet} 
\usepackage{courier}  
\usepackage[hyphens]{url}  
\usepackage{graphicx} 
\usepackage{natbib}  
\usepackage{caption} 
\title{Intelligent Frame Selection as a  Privacy-Friendlier Alternative to Face Recognition
}

\author{
    Mattijs Baert,
    Sam Leroux,
    Pieter Simoens
    \\
}
\affiliations{
  IDLab, Department of Information Technology, Ghent University - imec\\
    Technologiepark-Zwijnaarde 126\\
    B-9052 Gent, Belgium\\
    mattijs.baert@ugent.be, sam.leroux@ugent.be, pieter.simoens@ugent.be

}

\begin{document}

\maketitle

\begin{abstract}
The widespread deployment of surveillance cameras for facial recognition gives rise to many privacy concerns. This study proposes a privacy-friendly alternative to large scale facial recognition. While there are multiple techniques to preserve privacy, our work is based on the minimization principle which implies minimizing the amount of collected personal data. Instead of running facial recognition software on all video data, we propose to automatically extract a high quality snapshot of each detected person without revealing his or her identity. This snapshot is then encrypted and access is only granted after legal authorization. We introduce a novel unsupervised face image quality assessment method which is used to select the high quality snapshots. For this, we train a variational autoencoder on high quality face images from a publicly available dataset and use the reconstruction probability as a metric to estimate the quality of each face crop. We experimentally  confirm that the reconstruction probability can be used as biometric quality predictor. Unlike most previous studies, we do not rely on a manually defined face quality metric as everything is learned from data. Our face quality assessment method outperforms supervised, unsupervised and general image quality assessment methods on the task of improving face verification performance by rejecting low quality images. The effectiveness of the whole system is validated qualitatively on still images and videos.
\end{abstract}

\section{Introduction}
Recent advances in computer vision and machine learning have dramatically increased the accuracy of face recognition technologies \cite{schroff2015facenet, masi2018deep, deng2019arcface}. Face recognition is already commonly used in commercial products such as Apple's FaceID \cite{faceid} or at border checkpoints in airports where the portrait on a digitized biometric passport is compared with the holder's face. Most people have little to no concerns about these specific applications as they are limited in scope and have a single well defined goal. As the technologies mature however it becomes possible to deploy them on a much larger scale. The most well known example of this is the large scale use of CCTV cameras equipped with intelligent analysis software. In this way, facial recognition checkpoints are deployed at areas like gas stations, shopping centers, and mosque entrances \cite{larson2018china, NOS}.
\begin{figure}
    \centering
    \includegraphics[width=\linewidth]{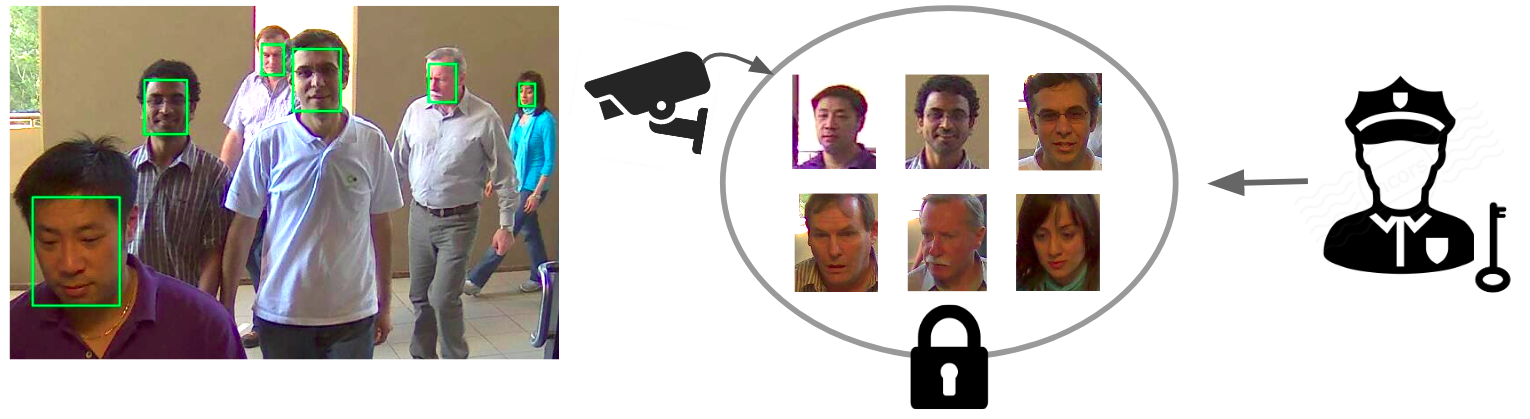}
    \setlength{\abovecaptionskip}{-2pt}
    \setlength{\belowcaptionskip}{-2pt}
    \caption{Overview of our proposed approach. For each subject, we automatically extract one high quality frame (or face crop) and encrypt it for storage. Access to the images is only possible after legal authorization in case of an investigation.}
    \label{fig:overview}
\end{figure}
From a domestic security point of view, these technologies are extremely useful. They can be used to find missing children, identify and track criminals or locate potential witnesses. There are many examples where CCTV footage in combination with facial recognition software has supported and accelerated criminal investigations. In 2016 for example, the ``man in the hat'' responsible for the Brussels terror attacks was identified thanks to FBI facial recognition software \cite{maninhat}. 
\\
\newline
The proliferation of facial recognition technology however also raises many valid privacy concerns. A fundamental human rights principle is that surveillance should be necessary and proportionate. This principle was adopted by the UN Human Rights Council (HRC) in the ``the right to privacy in the digital age'' resolution which states that ``States should ensure that any interference with the right to privacy is consistent with the principles of legality, necessity and proportionality''\cite{unitednations}.
\\
\newline
Governments have to balance both aspects of this technology before they implement a certain solution. On the one hand, there are the obvious advantages of a large scale blanket surveillance system but this clearly violates the proportionality principle. On the other hand, a recent study showed that a majority of the public considers it to be acceptable for law enforcement to use facial recognition tools to assess security threats in public spaces as long as it is within a clearly defined regulatory framework \cite{smith2019more}.
\\
An interesting research direction is the development of more privacy-friendly alternatives that can still support law enforcement in criminal investigations. In this work we present an intelligent frame selection approach that can be used as a building block in a more privacy-friendly alternative to large scale face recognition. The problem of key frame selection is defined as selecting a single frame out of a video stream that represents the content of the scene \cite{wolf1996key}. In the context of facial recognition, our goal is to record a single clear crop of each person visible in the stream without revealing his or her identity. In such a way, the minimization strategy of privacy preserving technologies \cite{DomingoFerrer2020} is implemented by collecting the minimal necessary amount of personal data.
According to \cite{duncan2007engineering}, data minimization is an unavoidable first step to engineer systems in line with the principles of privacy by design \cite{cavoukian2009privacy}. Next, to ensure the confidentiality of the collected data, all images are encrypted (i.e. ``hide'' strategy) and access can only be provided to law enforcement after legal authorization as shown in figure \ref{fig:overview}.
\\
\newline
Extracting face crops which are suitable for recognition is a difficult problem as surveillance footage is typically blurry because of the motion of the subjects. In addition, it is hard to quantify the quality of a single frame as it depends on multiple aspects such as the angle, illumination and head position. Frame selection techniques have been used before in the context of face recognition  to reduce the computational cost or increase the recognition rate \cite{wong_cvprw_2011, anantharajah2012quality, qi2018boosting, vignesh2015face, best2017automatic, hernandez2020biometric, terhorst2020ser}. In this work we introduce a novel deep learning based technique to perform intelligent face recognition. The key contribution is a new face image quality assessment (FQA) approach inspired by anomaly detection. Unlike previous solutions, our system is trained completely unsupervised without the need for labeled face images and without access to a face recognition system. Our approach can be used for the same tasks as previous frame selection techniques but in this work we propose to use it as a building block for a more privacy-friendly alternative to large scale face recognition.
\\
\newline
This paper is organized as follows. An overview of the related work is presented in section \ref{section:related}. Next, we propose a privacy preserving alternative to large scale facial recognition in section \ref{section:privacy_preserving_appr}. Our novel FQA method is introduced in section \ref{section:approach} and experimentally validated in  section \ref{section:experimental_setup} and \ref{section:results}. We conclude in section \ref{section:conclusion} and give some pointers for future research.

\section{Related work}
\label{section:related}
Deep learning has dominated the field of facial recognition in the past years. Recent techniques can reach accuracies of 99.63\% \cite{schroff2015facenet} on the Labeled Faces in The Wild dataset \cite{huang2008labeled}, the most commonly used benchmark dataset. These data-driven methods outperform techniques based on engineered features by a large margin \cite{masi2018deep}. Face recognition is typically subdivided into face verification and face identification. Face verification is the task of deciding whether two pictures show the same person where face identification tries to determine the identity of the person on an image.
\\
\newline
A large amount of research is dedicated to extracting high quality face image representations. These can then be used to calculate the similarity between two face crops or as input to a classification model. Different loss functions have been designed to get meaningful face image representations. First attempts used softmax loss \cite{sun2014deep} while recent approaches focus on euclidean distance-based loss \cite{schroff2015facenet}, cosine-margin-based loss \cite{wang2017normface} and variations of the softmax loss \cite{deng2019arcface}. Given these high quality feature embeddings, we can directly perform face verification by calculating the distance between two images (i.e. one-to-one matching). Face identification requires a database which contains a reference image of every identity. To identify a person on an image, the probe image's embedding should be compared with all images in the reference database (i.e. one-to-many matching).
\\
\newline
Face images are a popular biometric since they can be collected in unconstrained environments and without the user's active participation. These properties in combination with the widespread deployment of surveillance cameras give rise to some severe privacy concerns.  As a result, researchers explore techniques to secure the data collected by surveillance cameras and are developing privacy preserving facial recognition systems. The techniques differ in exactly what privacy-sensitive aspects they protect. Some techniques avoid deducing soft biometrics (e.g. age, gender, race) from data collected for verification or recognition purposes \cite{mirjalili2018gender, mirjalili2020privacynet}. Other techniques focus on face image de-identification \cite{newton2005preserving}, which eliminates the possibility of a facial recognition system to identify the subject while still preserving some facial characteristics. A very powerful technique hides the input image and the facial recognition result from the server that performs the recognition using a cryptographic enhanced facial recogniton system \cite{erkin2009privacy}.
\\
\newline
A lot of effort has gone into the development of face image quality assessment (FQA) metrics. Face image quality is defined as the suitability of an image for consistent, accurate and reliable recognition \cite{hernandez2020biometric}. FQA methods aim to predict a value which describes the quality of a probe image. Most previous FQA techniques focus on improving the performance of a face recognition system \cite{wong_cvprw_2011, anantharajah2012quality, qi2018boosting, vignesh2015face, best2017automatic, hernandez2020biometric, terhorst2020ser}. An FQA system is then used as a first step to make sure that we only feed high quality images to the face recognition model, therefore increasing the accuracy and reducing the computational cost. A first family of FQA techniques (Full-reference and reduced-reference FQA) assumes the presence of a high quality sample of the probe image's subject. These methods do not work for unknown subjects, which is necessary for our purpose as we will explain in the next section. A second family of FQA techniques develops hand-crafted features to assess the quality of a face image \cite{anantharajah2012quality, REFICAO} while more recent studies apply data driven methods and report a considerable increase in performance. Different studies \cite{qi2018boosting, vignesh2015face} propose a supervised approach where a model is trained to predict the distance between the feature embeddings of two images. Since two samples are necessary to perform a comparison, one low quality sample can affect the quality score of a high quality sample. This is commonly solved by assuming that an image compliant with the ICAO guidelines \cite{REFICAO} represents perfect quality \cite{hernandez2020biometric}. In the work of Hernandez-Ortega et al. a pretrained resnet-50 network \cite{he2016deep} is modified by replacing the classification layer with two regression layers which output the quality score. Alternatively, it is also possible to use human labeled data \cite{best2017automatic}. The most similar to our work is \cite{terhorst2020ser} which also proposes an unsupervised approach. Here the quality of an image is measured as its robustness in the embedding space, which is calculated by generating embeddings using random subnetworks of a selected face recognition model.
\\
\newline
Compared to previous work, we introduce a novel completely unsupervised FQA method based on a variational autoencoder. We assume no access to the identities of the people and show that it works well for unseen people. Unlike \cite{terhorst2020ser}, no facial recognition model is used to calculate a quality score. In contrast to previous work our main goal is not necessarily to improve the facial recognition accuracy but instead we use it as a building block to enable a more privacy-friendly alternative to large scale face recognition, as explained in the next section.

\section{Frame selection as an alternative to face recognition}
\label{section:privacy_preserving_appr}
In this section we introduce a framework based on \cite{simoens2013scalable} that uses intelligent frame selection in the context of face recognition to build a more privacy-friendly alternative to large scale face recognition. Instead of proactively trying to recognize individuals, we follow the presumption of innocence principle and do not indiscriminately perform recognition of people as they go about their daily business. Instead, our system uses key frame extraction to capture a high quality snapshot of every person passing by the camera. These snapshots are encrypted and stored locally on the camera or securely transmitted to a cloud back-end for storage. In a normal situation this data is then automatically deleted after a well defined time period as defined in article 17 of the GDPR (``Right to be forgotten''). In this case, the identity of the people will never be known and no human operator will have access to the unencrypted pictures. The images can only be decrypted after legal authorization in case of a criminal investigation or another situation where access to the identities present in a certain location at a certain time is warranted. The high quality images can then be used as input for a face recognition system or to aid the investigation process. Since only high quality crops are stored, the storage overhead is much lower than in a system where the full video is stored. This system is summarized in Figure \ref{fig:overview} using a video from the ChokePoint dataset \cite{wong_cvprw_2011}.
\\
\newline
It is important to note that we need to store at least one crop for each person visible in the video. It is not enough to use a fixed threshold to decide whether a frame should be stored or not, instead we have to store the best frame for each individual even if this best frame still has a relatively low quality compared to images of other individuals (for example because the person never perfectly faces the camera). As a generalization we could also decide to store a short video clip of a few seconds before and after the best frame has been captured.
\\
\newline
An obvious disadvantage of our approach is that it is not possible to proactively recognize people for example to detect wanted individuals. On the other hand it does support the criminal investigation after the fact. Our system is therefore complementary and more suited to low risk areas where a full blown face recognition system would violate the proportionality principle.

\section{Face image quality assessment}
\label{section:approach}
\begin{figure}[t]
    \centering
    \includegraphics[width=\linewidth]{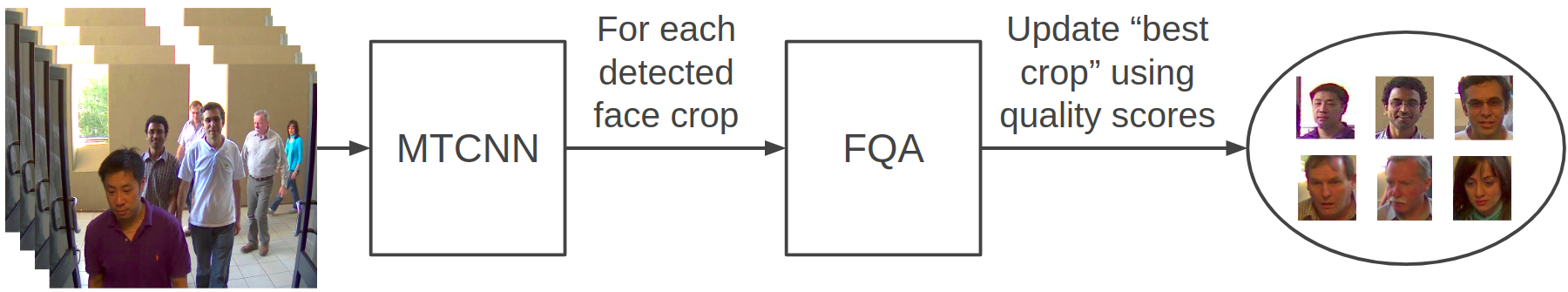}
    \caption{Overview of our face quality-aware frame selection system.}
    \label{fig:fqa_overview}
\end{figure}
\begin{figure}
    \centering
    \includegraphics[width=225pt]{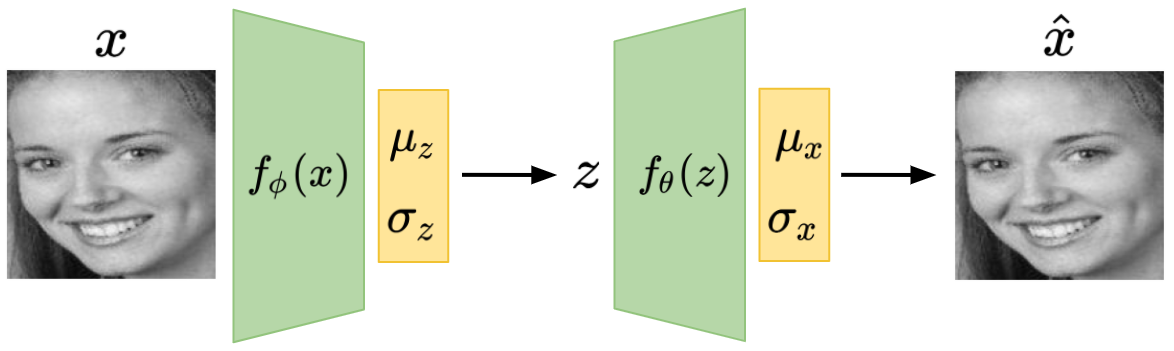}
    \caption{Variational autoencoder with encoder $f_{\phi}(x)$ and decoder $f_{\theta}(z)$, each arrow represents a sampling processes.}
    \label{fig:vae}
\end{figure}

The system proposed in the previous section relies on a Face Quality Assessment block to decide which crops to store. Any FQA method can be used but in this section we introduce a novel technique based on a variational autoencoder. Compared to other FQA methods, this has the benefit of being completely unsupervised. We do not assume access to a face recognition system or to the identities of the people in the dataset. Our method also generalizes well to other people outside of the original training dataset.
\\
\newline
 An overview of the proposed system is depicted in figure \ref{fig:fqa_overview}. The first step is to detect all faces in each frame and track them across subsequent frames. We use the MTCNN model \cite{zhang2016joint} to detect faces in a still video frame. The output is a list of bounding boxes around the detected faces. To track a subject across a video, we simply calculate the euclidean distance between the bounding boxes of subsequent frames. Bounding boxes that are close to each other are considered to correspond to the same subject. To evaluate the quality of a face crop, we calculate the reconstruction probability of a variational autoencoder (VAE) trained on a dataset of high quality images. The VAE reconstruction probability is commonly used as an anomaly detection metric \cite{an2015variational} to determine how different an input is to the data seen during training. By training the VAE on a publicly available dataset of high quality face images, we reformulate the FQA task as an anomaly detection task (i.e. how different is this face crop from the high quality face crops seen during training?). The next paragraph explains the VAE and reconstruction probability in more details.
\\
\newline
A variational autoencoder (VAE) \cite{kingma2013auto} is a probabilistic variant of the standard autoencoder (AE). The encoder and decoder are modeled by probabilistic distributions rather than deterministic functions. The encoder $f_{\phi}(x)$ models the posterior $q_{\phi}(z | x)$ of the latent variable $z$ and a decoder $f_{\theta}(z)$ models the likelihood $p_{\theta}(x | z)$ of the data $x$ given the latent variable $z$. The prior distribution of the latent variable $p_{\theta}(z)$ is chosen as a Gaussian distribution $\mathcal{N}(0, I)$. The posterior $q_{\phi}(z | x)$ and likelihood $p_{\theta}(x | z)$ are isotropic multivariate normal distributions $\mathcal{N}(\mu_{z}, \sigma_{z})$ and $\mathcal{N}(\mu_{x}, \sigma_{x})$ respectively. Figure \ref{fig:vae} shows the process of a forward pass of an image $x$ through the VAE, the arrows represent a sampling process. To train a VAE using backpropagation, every operation should be differentiable which is not the case for the sampling operations: $z \sim \mathcal{N}(\mu_{z}, \sigma_{z})$ and $\hat{x} \sim \mathcal{N}(\mu_{x}, \sigma_{x})$. Applying the re-parameterization trick fixes this problem. A dedicated random variable $\epsilon \sim \mathcal{N}(0, 1)$ is sampled such that the sampling operations can be rewritten as: $z \sim \mu_{z} + \epsilon \cdot \sigma_z$ and $\hat{x} \sim \mu_{x} + \epsilon \cdot \sigma_x$. The VAE training objective is written as the expected log likelihood minus the KL divergence between the posterior and the prior as described in equation \ref{eq:train_obj}. 
\begin{equation}
    \mathcal{L}(x) = E_{q_{\phi}(z|x)}(p_{\theta}(x|z)) - KL(q_{\phi}(z|x)|p_{\theta}(z))
    \label{eq:train_obj}
\end{equation}
The first term is the reconstruction term and forces a good reconstruction $\hat{x}$ of the input data $x$. The KL term regularizes the distribution of the latent space by forcing it to be Gaussian. By training a generative model, like a VAE, the model learns to approximate the training data distribution. When a large reconstruction error is observed, this is an indication that the data is not sampled from the data distribution the VAE was trained on.
\\
\newline
The reconstruction probability is a generalization of the reconstruction error by taking the variability of the latent space and the reconstruction into account \cite{an2015variational}. First, an image $x$ is fed to the encoder which generates the mean vector $\mu_z$ and standard deviation vector $\sigma_z$. Next, $L$ samples $\{z^0, z^1,...,z^l\}$ are drawn from the latent distribution $\mathcal{N}(\mu_z,\sigma_z)$. All samples $z^l$ are fed into the decoder to get the distribution of the reconstruction of $x$ which is described by the mean $\mu^{l}_{\hat{x}}$ and the standard deviation $\sigma^{l}_{\hat{x}}$. The reconstruction probability is the probability of $x$ averaged over L samples as described by equation \ref{eq:r_prop}.
\begin{equation}
    \mathrm{RP}(x) = \frac{1}{L} \sum^{L}_{l=1} \mathcal{N}(x | \mu^{l}_{\hat{x}}, \sigma^{l}_{\hat{x}})
    \label{eq:r_prop}
\end{equation}
\newline
The reconstruction probability was originally developed as an anomaly score by \cite{an2015variational}. When a VAE is trained solely on samples of ``normal'' data, the latent distribution learns to represent these samples in a low dimensional space. Accordingly samples from ``normal'' data result in a high reconstruction probability while anomalies result in low reconstructing probabilities. We define the biometric quality of a face image as the reconstruction probability calculated by a VAE trained on high quality face images. Correspondingly, a high reconstruction probability is expected for high quality face images. Note that there is no explicit definition of face image quality and the quality score is independent of any face recognition model. The definition of a high quality face image is derived directly from the training data. 
\\
\newline
The encoder $f_{\phi}(x)$ consists of 5 consecutive blocks of a convolution layer, batch normalization and a leaky ReLU activation function with at the end two fully connected layers. The outputs of the encoder are the parameters defining $q_{\phi}(z|x)$. The decoder $f_{\theta}(z)$ consists of 5 blocks of a transposed convolution layer, batch normalization and a leaky ReLU activation function with again two fully connected layers at the end. The outputs of the decoder are the parameters defining $p_{\theta}(x|z)$. To calculate the reconstruction probability, $L$ is set to 10. The CelebA dataset \cite{liu2015faceattributes} consisting of 202,599 face images serves as training data. The Adam  optimization algorithm \cite{kingma2014adam} was applied with a learning rate of 0.005 and a batch size of 144. The VAE was trained for 10 epochs. 
Each image is cropped by MTCNN \cite{zhang2016joint}, resized to 64x64 pixels and converted to grayscale.

\section{Experimental setup}
\label{section:experimental_setup}
In this section, we isolate the FQA block for evaluation. According to the national institute of standards and technology (NIST), the default way to quantitatively evaluate a FQA system is analyzing the error vs. reject curve (ERC) \cite{grother2007performance, grother2020ongoing}. As defined in section \ref{section:related}, FQA indicates the suitability of an image for recognition. The ERC measures to what extent the rejection of low quality samples increases the verification performance as measured by the false non-match rate (FNMR). The FNMR is the rate at which a biometric matcher miscategorizes two signals from the same individual as being from different individuals \cite{Schuckers2010}. Face verification consists of calculating a comparison score of two images and comparing this score with some threshold. The comparison score is defined as the euclidean distance between the FaceNet \cite{schroff2015facenet} embeddings of the two images. To avoid a low quality sample affecting the verification performance of a high quality sample, one high quality reference image for every subject is required. For this, an ICAO compliant image is typically used \cite{hernandez2020biometric}. We used the BioLab framework \cite{ferrara2012face} to calculate an average ICAO compliance score for all images. For every subject,  the image with the highest ICAO score is selected as a reference image. Note that it is not possible to use the BioLab framework as face quality assessment method directly because it cannot assess all images accurately and it is unable to operate in real-time. Now, assume a set of genuine image pairs $(x_i^{(1)}, x_i^{(2)})$ (i.e. two images of the same person) of length N. Every image pair $(x_i^{(1)}, x_i^{(2)})$ constitutes a distance $d_i$ (i.e. comparison score). To determine if two images match, the distance between the two images is compared with a threshold $d_t$. Using quality predictor function $F$ (i.e. a FQA method), a quality value $q_i$ is calculated for each image pair. Since $x_i^{(1)}$ is always a high quality image from the reference database, the quality $q_i$ of image pair $(x_i^{(1)}, x_i^{(2)})$ can be written as:
\begin{figure}[t]
    \centering
    \includegraphics[width=225pt]{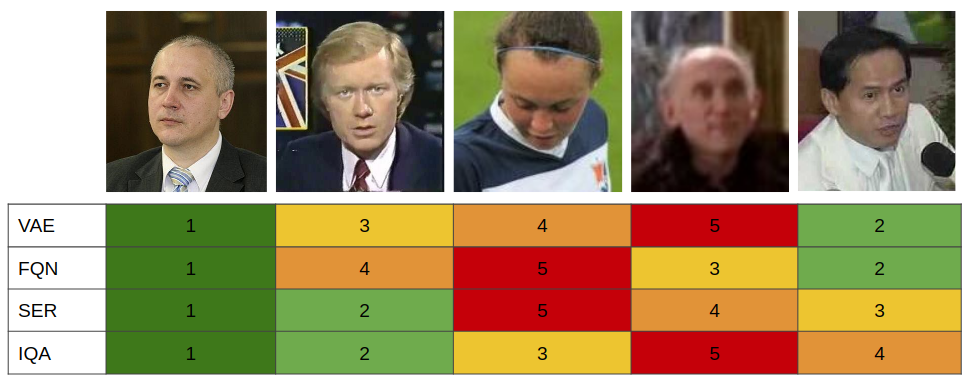}
    \caption{Sample images from the VggFace2 dataset ranked by different quality metrics.}
    \label{fig:example_images}
\end{figure}
\begin{equation}
    q_i = q_i^{(2)} = F(x_i^{(2)}).
\end{equation}
Now consider $R$ as a set of low quality entities composed by the samples that correspond with a predicted quality value below some threshold:
\begin{equation}
    R(r) = \{i : q_i < F^{-1}(r),  \forall i < N \}.
\end{equation}
$F^{-1}$ is the inverse of the empirical cumulative distribution function of the predicted quality scores. The parameter $r$ is the fraction of images to discard, such that $F^{-1}(r)$ equals the quality threshold that corresponds with rejecting a fraction $r$ of all images. Then the FNMR can be written as:
\begin{equation}
    \mathrm{FNMR} = \frac{|d_i : d_i \geq d_t, i \notin R(r)|}{|d_i : d_i \geq -\infty, i \notin R(r)|}
\end{equation}
The value of $r$ is manipulated to calculate the FMNR for different fractions of rejected images. The value of $d_t$ is fixed and computed using the inverse of the empirical cumulative distribution function of the distances between reference and probe images $M^{-1}$:
\begin{equation}
    d_t = M^{-1}(1 - f).
\end{equation}
Practically, $f$ is chosen to give some reasonable FNMR. As suggested by \cite{frontex} a maximum FNMR of 0.05 is maintained.
\section{Results}
\label{section:results}
\subsection{VggFace2}
The VggFace2 dataset \cite{Cao18} is designed for training face recognition models and consists of more than 3.3 million face images of more than 9000 identities. In the following experiments, our approach (VAE) is compared to FaceQNet (FQN) \cite{hernandez2020biometric}, a method based on the stochastic embedding robustness (SER) \cite{terhorst2020ser} and a general image quality assessment (IQA) system \cite{talebi2018nima}. FaceQNet is a fine-tuned resnet-50 network \cite{he2016deep} which is trained on large amounts of face images for a face recognition task. The classification layer is then replaced by two layers designed for quality regression. The ground truth quality score is defined as the euclidean distance between the feature embeddings of the probe image and an ICAO compliant image of the same subject. The face image quality assessment method based on stochastic embedding robustness (SER) calculates a quality score by measuring the variations of the embeddings generated by random subnetworks of a resnet-101 \cite{he2016deep} model trained for facial recognition. The general image quality assessment (IQA) system \cite{talebi2018nima} does not take face features into account and predicts the general image quality. For all conducted experiments, images were cropped by the MTCNN face detector \cite{zhang2016joint}.
\\
\newline
Figure \ref{fig:example_images} shows five images from the VggFace2 dataset \cite{Cao18} ranked by different quality metrics. This allows a first qualitative evaluation of the five metrics. As presented on the figure, all metrics agree on assigning the highest quality to the first image. All FQA metrics assign a low quality value to the third image because the person looks down, the general IQA method does not take this aspect into account and assigns a higher quality value. For the same reason, the IQA method assigns a low quality value to the fifth image in contrast to all FQA algorithms. On the fourth image, our method agrees with the IQA method and selects it as worst quality image as opposed to the other FQA metrics. We could assume that our method attaches more importance to blurriness compared to the other FQA algorithms. It is remarkable that the second image is ranked as second worst by FQN since the person looks straight into the camera.
\begin{figure}[t]
    \centering
    \includegraphics[width=225pt]{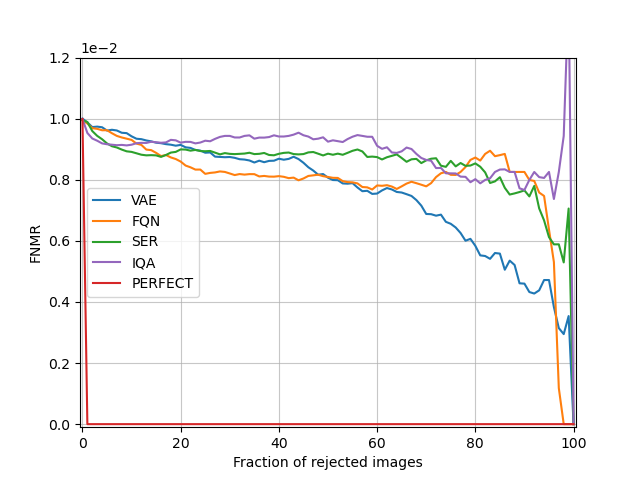}
    \caption{ERC with an initial FNMR of 0.01.}
    \label{fig:vggface2_fnmr_01}
\end{figure}
\begin{figure}[h]
    \centering
    \includegraphics[width=225pt]{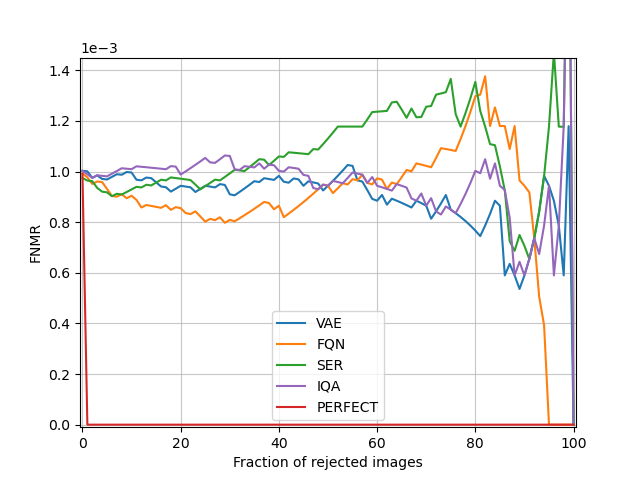}
    \caption{ERC with an initial FNMR of 0.001.}
    \label{fig:vggface2_fnmr_001}
\end{figure}
\\
\newline
In a second experiment, we evaluate our proposed FQA algorithm on still face images by analyzing the ERC plots as explained in section \ref{section:experimental_setup}. For 103.008 images of 300 subjects from the VggFace2 dataset \cite{Cao18}, a high quality reference image is selected using the ICAO compliance scores. The ERC plots in figures \ref{fig:vggface2_fnmr_01} and \ref{fig:vggface2_fnmr_001} display the FNMR for different fractions of rejected images. The red line with the label ``PERFECT'' represents a quality measure which correlates perfectly with the distance scores. When an initial FNMR of 0.01 is set, in an ideal scenario, the FNMR will be zero after rejecting 1\% of all images. The closer an ERC is to the red line, the better the performance of the used FQA algorithm. For an initial FNMR of 0.01 our approach clearly outperforms FaceQNet, SER and the general image quality assessment program. We hypothesize that SER would perform better when the same type of embeddings were used for verification as quality estimation. In the conducted experiments SER uses ArcFace \cite{deng2018arcface} embeddings to estimate face image quality while the FNMR is calculated using FaceNet embeddings. For an initial FNMR of 0.001, the difference with the other approaches is smaller. It is important to note that our model is considerably smaller than FaceQNet and the resnet-101 model \cite{he2016deep} used by SER.
\begin{figure}[h!]
    \centering
    \includegraphics[width=225pt]{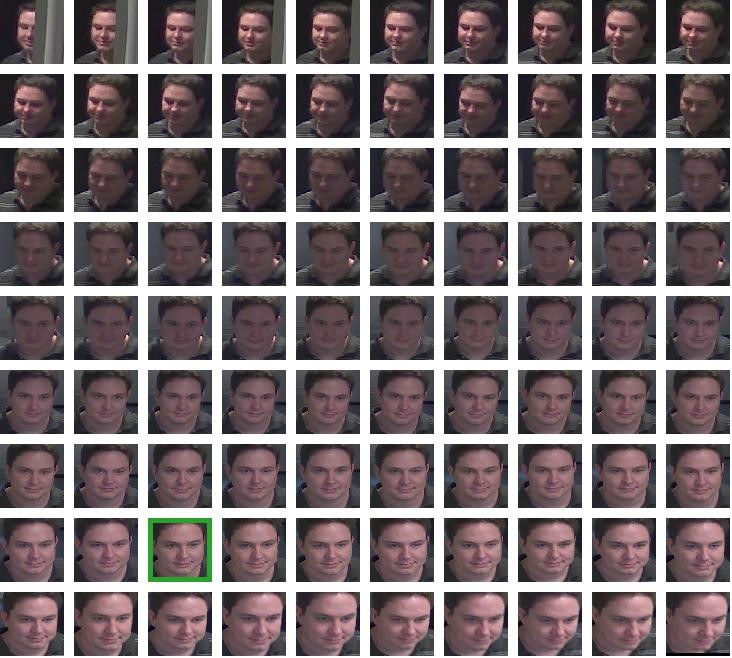}
    \caption{Consecutive face crops from one tracked identity in the ChokePoint dataset. The crop with the green border corresponds with the highest quality calculated by our FQA algorithm.}
    \label{fig:cp_map_0}
\end{figure}
\begin{figure}[h!]
    \centering
    \includegraphics[width=225pt]{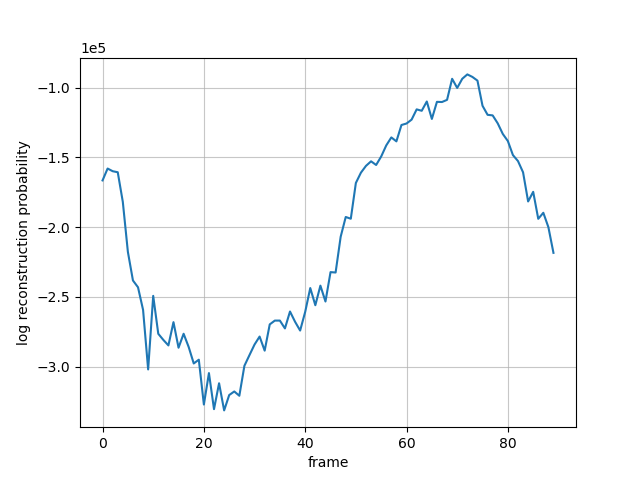}
    \caption{Logarithm of the reconstruction probability (i.e. face quality) for consecutive face crops.}
    \label{fig:cp_plot_0}
\end{figure}
FaceQNet comprises 7 times more parameters than our VAE and resnet-101 even 14 times more. Additionally, our method is trained completely unsupervised without the need for ground truth quality values while FaceQNet relies on distance scores as ground truth quality values. The ground truth generation process used by FaceQNet also indicates the dependency on one or more face recognition models. This dependency is even more prominent for SER since a specific face recognition network is used for generating the quality predictions.
\subsection{ChokePoint}
In a third experiment, we focus on the ChokePoint dataset \cite{wong_cvprw_2011}. This dataset is designed for conducting experiments on face identification and verification under real-world surveillance conditions. The dataset consists of 48 videos or 64,204 face images. Both crowded and single-identity videos are included. We now evaluate our system qualitatively on selecting one high quality face crop of each identity in a video stream. Figure \ref{fig:cp_map_0} shows consecutive face crops of an example video. The crop outlined in green is the frame that corresponds with the highest quality value calculated by our FQA algorithm. Figure \ref{fig:cp_plot_0} shows how the quality score changes over time as the subject moves through the scene. We define the quality score as the logarithm of the reconstruction probability. We can see that initialy the quality score decreases as the person is moving through a darker area looking down. The shades and the angle of the face makes these crops less useful for face recognition. As the person moves closer to the camera, the brightness increases and the subject becomes clearly visible. This is also reflected in an increasing quality score. The highest score is obtained when the person is close to the camera and is looking in the general direction of the camera. As the person turns away, the score again decreases. This qualitative example shows that our model is indeed able to assign understandable and meaningful scores to each frame. We made videos of this and other examples publicly available \footnote{\url{https://drive.google.com/drive/folders/1GRlFRSxHRfBnfTpI5DG2v2rN3nTCg5Y0?usp=sharing}}.   
\subsection{Bicycle parking}
\begin{figure}
    \centering
    \includegraphics[width=225pt]{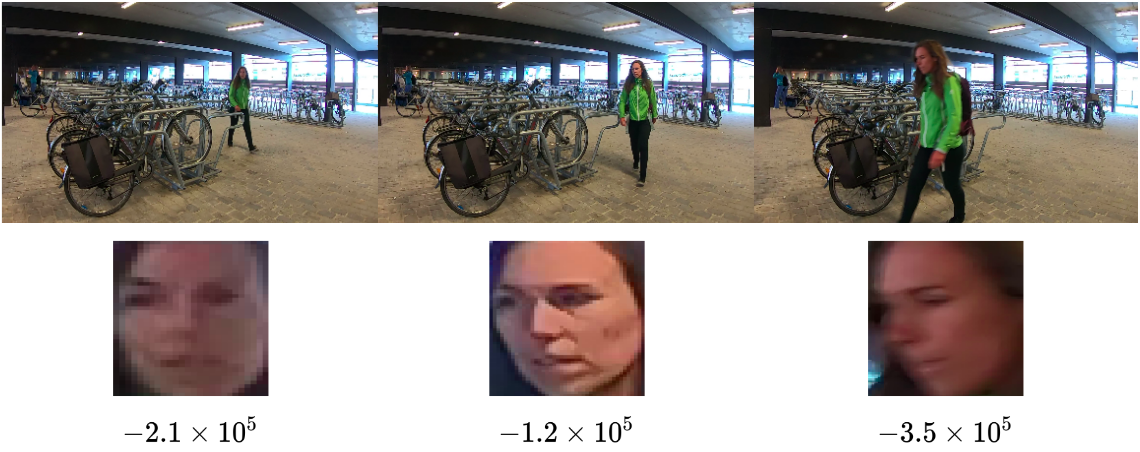}
    \caption{Three frames from the footage of a security camera in a bicycle parking lot. The corresponding frame crops and quality scores are depicted below each frame.}
    \label{fig:bicycle_park}
\end{figure}

Finally, we also validated our approach on video data from security cameras in a bicycle parking lot. This to investigate how well the model generalizes to data collected in the real world. Figure \ref{fig:bicycle_park} shows three  example frames with its corresponding frame crop and quality score. Even though these images are very different from the images the VAE was originaly trained on, we can see that the model generalizes well and is able to assign useful scores to each crop.

\section{Conclusion and future work}
\label{section:conclusion}
In this study, a novel face image quality assessment method is proposed based on a variational autoencoder's reconstruction probability. This is, by our knowledge, the first time a generative model like a VAE is used to tackle the problem of face image quality assessment. We demonstrate, by quantitative and qualitative results, that our method can be used as a biometric quality predictor. Unlike other data driven approaches, no facial recognition model is used for training and no explicit definition of face quality is given. Our FQA algorithm is used as a building block in a privacy-friendly alternative to large scale facial recognition. Instead of identifying all detected faces in a video stream, our system saves one high quality face crop without revealing the person's identity. This face crop is encrypted and access is only granted after legal authorization. In such a way, the system still supports criminal investigations while not violating the proportionality principle.
\\
\newline
In future work, we will further optimize the VAE architecture keeping the constraints on model size and computational complexity in mind as the final goal would be to deploy the model on a stand-alone edge device. It would be interesting to investigate different hardware platforms such as FPGAs that allow the model to process data in real-time with a small energy consumption, making it possible to embed our system in low cost surveillance camera's. Moreover, our method should be evaluated on other datasets and in combination with alternative feature extractors.

\section{ Acknowledgments}
We thank Klaas Bombeke (MICT UGent) for providing us with the dataset used in the experiment of Figure~9, and Stijn Rammeloo from Barco for the feedback on the initial manuscript. This research was funded by the imec.ICON SenseCity project and the Flemish Government (AI Research Program).

\bigskip

\bibliography{Main}

\newpage

\end{document}